\documentclass[letterpaper]{article} 
\usepackage{aaai25}  
\usepackage{times}  
\usepackage{helvet}  
\usepackage{courier}  
\usepackage[hyphens]{url}  
\usepackage{graphicx} 
\usepackage{natbib}  
\usepackage{caption} 
\usepackage{algorithm}
\usepackage{algorithmic}
\usepackage{xcolor}
\usepackage{float}

\usepackage{amsmath}
\usepackage{amssymb}
\usepackage{multirow}
\usepackage{graphicx}
\newcommand{\ie}{\textit{i}.\textit{e}.}
\newcommand{\eg}{\textit{e}.\textit{g}.}
\usepackage{enumerate}

\usepackage{newfloat}
\usepackage{listings}
\usepackage{threeparttable} 
\usepackage{multirow}
\usepackage{makecell}
\usepackage{array}
\usepackage{algorithm}
\usepackage{algorithmic}
\usepackage{color}
\usepackage{amsmath}
\usepackage{amsfonts}
\usepackage{threeparttable} 
\usepackage{booktabs}
\usepackage{multirow}
\usepackage{makecell}
\usepackage{siunitx}
\usepackage{bicaption}
\DeclareCaptionStyle{ruled}{labelfont=normalfont,labelsep=colon,strut=off} 
\floatstyle{ruled}
\newfloat{listing}{tb}{lst}{}
\floatname{listing}{Listing}
\nocopyright

\title{A Training-free Synthetic Data Selection Method for Semantic Segmentation}

\author{
    Hao Tang\textsuperscript{\rm 1}, Siyue Yu\textsuperscript{\rm 2}, Jian Pang\textsuperscript{\rm 1}, Bingfeng Zhang\textsuperscript{\rm 1}\thanks{Corresponding author.}
}
\affiliations{
    \textsuperscript{\rm 1}China University of Petroleum (East China)\\ 
    \textsuperscript{\rm 2}XJTLU\\

    \{haotang, jianpang\}@s.upc.edu.cn, siyue.yu@xjtlu.edu.cn, Bingfeng.Zhang@upc.edu.cn
}

\usepackage{bibentry}

\begin{document}

\maketitle
\begin{abstract}
Training semantic segmenter with synthetic data has been attracting great attention due to its easy accessibility and huge quantities. Most previous methods focused on producing large-scale synthetic image-annotation samples and then training the segmenter with all of them. However, such a solution remains a main challenge in that the poor-quality samples are unavoidable, and using them to train the model will damage the training process. In this paper, we propose a training-free Synthetic Data Selection (SDS) strategy with CLIP to select high-quality samples for building a reliable synthetic dataset. Specifically, given massive synthetic image-annotation pairs, we first design a Perturbation-based CLIP Similarity (PCS) to measure the reliability of synthetic image, thus removing samples with low-quality images. Then we propose a class-balance Annotation Similarity Filter (ASF) by comparing the synthetic annotation with the response of CLIP to remove the samples related to low-quality annotations. The experimental results show that using our method significantly reduces the data size by half, while the trained segmenter achieves higher performance.

\begin{links}
\link{Code}{https://github.com/tanghao2000/SDS}
\end{links}

\end{abstract}

%

\section{Introduction}
Semantic segmentation is a fundamental task in computer vision~\cite{chen2017deeplab, he2017mask, liu2018path, zhao2017pyramid}. Its goal is to assign semantic labels to each pixel in an image, which is crucial for applications such as autonomous driving~\cite{liu2020importance}, semantic editing~\cite{ling2021editgan}, and medical image segmentation~\cite{ronneberger2015u}. 

\begin{figure}[!t]
\centerline{\includegraphics[width=1.0\columnwidth]{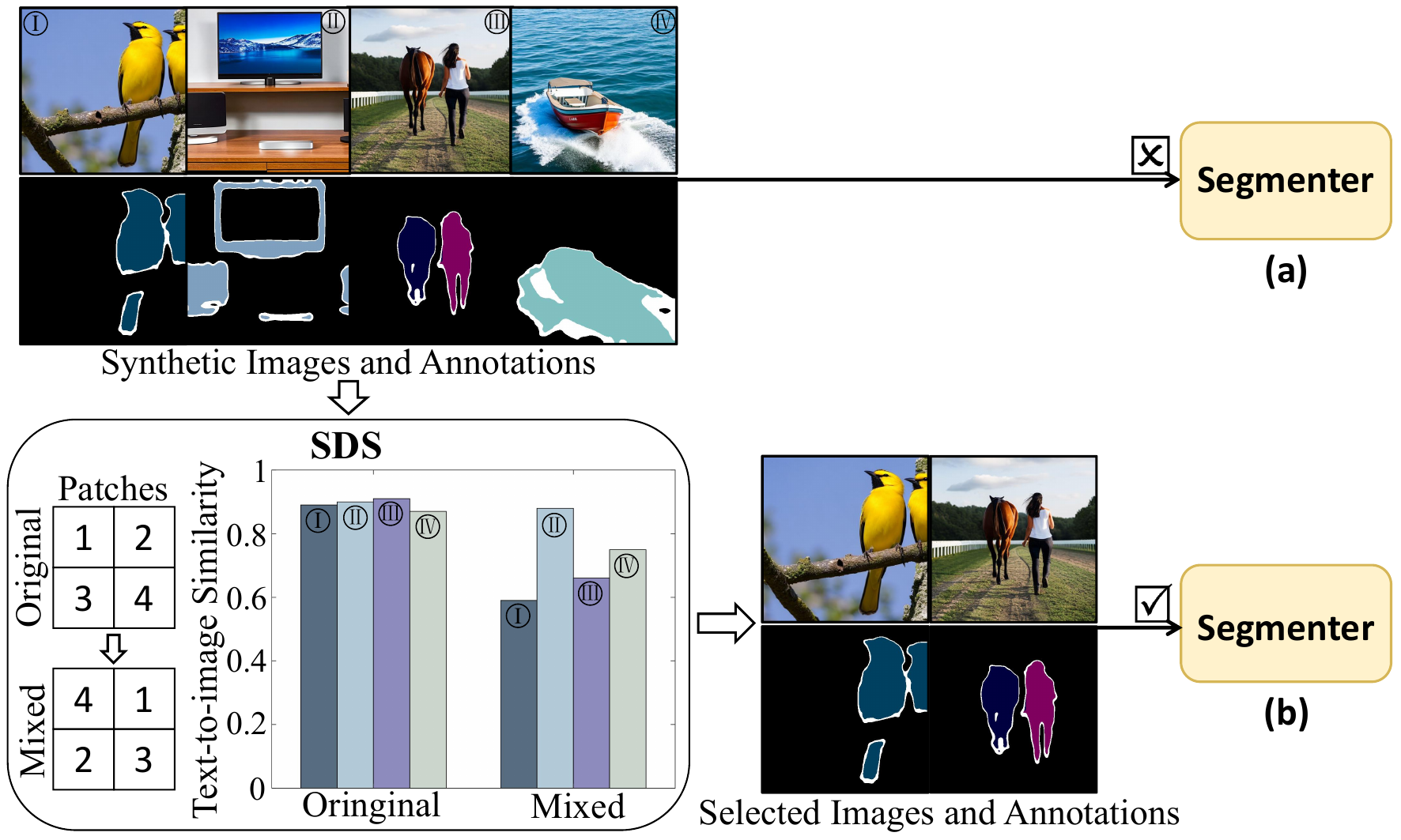}}
\caption{Comparison of synthetic data training methods. (a) Previous methods use all the synthetic data to train a segmenter. (b) Our training-free synthetic data selection method (SDS) selects higher-quality samples to train the segmented, where we select images with significant text-to-image similarity differences before and after mixed image patches.}
\label{intro}
\end{figure}

With the increasing demand for large-scale datasets in semantic segmentation tasks, the use of synthetic data has attracted widespread attention from researchers. Previous researchers utilize Generative Adversarial Networks (GANs)~\cite{creswell2018generative} and their variants, like DatasetGAN~\cite{zhang2021datasetgan} and BigDatasetGAN~\cite{li2022bigdatasetgan} to effectively generate synthetic dataset, thereby reducing manual annotation. In recent years, Denoising Diffusion Probabilistic Models (DDPMs)~\cite{ho2020denoising, rombach2022high} have achieved astonishing text-to-image synthesis ability. Such models are also utilized in sample generation for semantic segmentation and can be divided into two pipelines: mask-to-image~\cite{xue2023freestyle, yang2024freemask} and image-to-mask~\cite{wu2023diffumask, nguyen2024dataset}. The mask-to-image pipeline can ensure the accuracy of annotations, but time-consuming with manual annotations. Instead, the image-to-mask method does not need human annotation at all, providing a more efficient pipeline. For example, DiffuMask~\cite{wu2023diffumask} and Dataset-Diffusion~\cite{nguyen2024dataset}, which effectively generate abundant pairs of synthetic images and annotations. Specifically, DiffuMask \cite{wu2023diffumask} leverages the cross-attention mapping text to image and trains the Affinity Net~\cite{ahn2018learning} to extend text-driven image synthesis to semantic mask generation. Furthermore, Dataset-Diffusion \cite{nguyen2024dataset} makes innovations based on DiffuMask. It rewrites prompts by a large language mode (LLM), generating realistic images and simultaneously producing corresponding segmentation masks. Note that after generating massive samples, these methods use \textbf{all the samples} to train the segmenter.

However, it is hard to control the generation process of the synthetic samples (synthetic image-annotation pairs), making it inevitable to generate samples whose distribution or domain is different from real samples, \emph{i.e.}, low-quality samples. In this case, training the segmenter with them makes it easy to learn unreliable information, impeding the segmentation performance, as shown in Fig.~\ref{intro}(a). Therefore, if we can recognize and select high-quality samples, \emph{i.e.}, find images fitting the distribution of the real-world image, and with accurate synthetic annotation, the effectiveness of the whole training process would be better guaranteed.


Based on the above analysis, we propose a training-free Synthetic Data Selection (SDS) strategy with the Contrastive Language Image Pretraining (CLIP) model~\cite{radford2021learning} to select high-quality samples, as shown in Fig.~\ref{intro} (b). Our intuition is that CLIP is trained on a large amount of real data. In theory, it fits the distribution of real data, making it possible to distinguish whether the synthetic image~\cite{wang2023exploring} belong to the same distribution of real images. To evaluate it, We randomly select several synthetic images using the Dataset-Diffusion~\cite{nguyen2024dataset} and calculate the text-to-image similarity. As shown on the left bar chart in Fig.~\ref{intro} (b), we find that all of them generate high text-to-image similarities, which is hard to regard as a confidence metric directly. Then we mix up the order of the image patches~\cite{lee2024entropy} and re-calculate the similarities. The results show a significant difference among the images, as shown on the right bar chart in Fig.~\ref{intro}(b). Considering the CLIP model is pre-trained with the natural object order of images, using the image with mixed patches as input, the text-to-image similarity should be low since mixing operation damages object structures, the model cannot receive common object relationships or the proper object order in the image. On the contrary, if the text-to-image similarity of patch-mixed images remains high, it can be treated as that the CLIP model relies on uncommon or even incorrect object relationships, \ie, unrepresentative information in the image makes the network produce high-confidence decisions. Such samples are unreliable in training a model. Therefore, we claim that a high-quality image should have low text-to-image similarity after mixing patches, while a low-quality image should have high similarity after mixing patches.

Following our observations, we design a Perturbation-based CLIP Similarity (PCS) approach to select reliable synthetic data for training semantic segmenter. Specifically, we first calculate text-to-image similarities for the original image and the patch-mixed image, respectively. For a high-quality image, its text-to-image similarity with the original patches should be high to guarantee it has the correct classes and objects. Meanwhile, its text-to-image similarity with mixed patches should be low to ensure it shares a similar distribution with real images. To accurately quantify the similarity degree, we use the similarity difference between the original patches and mixed patches to replace only considering the similarity of the mixed patches. The similarity difference is defined as the PCS score in this paper. Only samples that satisfy the above two rules will remain for further processing.

Besides, it is necessary for a high-quality sample to require both high-quality images and annotations. The above module only selects high-quality images, ignoring the annotation quality, as shown in Fig.~\ref{intro}, the PCS score of the boat is high, but it mistakenly labeled the wave as the boat. To solve this problem, we propose a class-balance Annotation Similarity Filter (ASF) to remove low-quality annotation samples by comparing the synthetic annotation with the response of CLIP. While selecting the synthetic images, we utilize the generation ability of CLIP to generate a set of reference annotations. Considering that the quality of annotations varies among different classes, we classify them into different groups and finally select high-quality annotation samples by computing the mIoU between reference annotations and synthetic annotations.


To evaluate the effectiveness of our selection strategy, we use our approach to select samples from synthetic datasets to train the segmenter and then evaluate the performance on two real-world datasets, PASCAL VOC 2012~\cite{everingham2015pascal} and MS COCO 2017~\cite{lin2014microsoft}. Experimental results show our method achieves higher performance while significantly reducing the dataset scales.

In summary, the contributions of our work are as follows:
\begin{itemize}

\item We observe that CLIP has different performances on different samples in synthetic data. Based on this, we propose a training-free Synthetic Data Selection (SDS) pipeline that can effectively select synthetic samples.

\item We design a Perturbation-based CLIP Similarity (PCS) method based on CLIP to select high-quality synthetic images. In addition, we also propose a class-balance Annotation Similarity Filter (ASF) module that compares synthetic annotations with the response of CLIP to remove samples associated with low-quality annotations. 

\item The experiment shows that with our data selection pipeline, the number of datasets is significantly reduced, and better performance is achieved in training segmenter, \eg, the synthetic training dataset can be reduced by half but generate a 3\% mIoU increase to 62.5\% when evaluating on the Pascal VOC 2012 dataset.
\end{itemize}

\begin{figure*}[!t]
\centerline{\includegraphics[width=\textwidth]{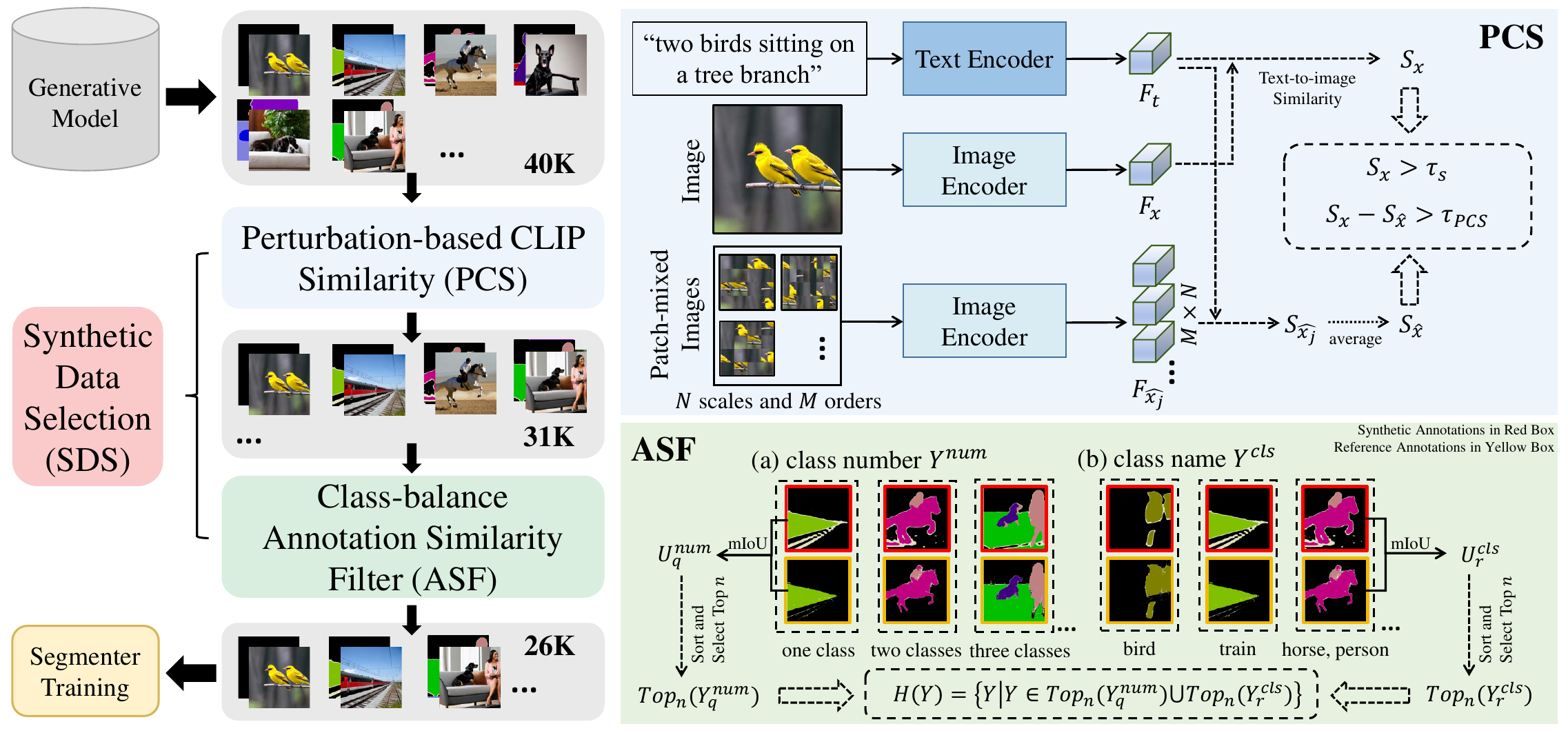}}
\caption{Overview of our Synthetic Data Selection (SDS) framework. (1) The initial synthetic dataset consists of 40k image-annotation pairs generated by a generative model. (2) Our SDS includes two modules: PCS and ASF. In the PCS module, the text caption and original image are encoded to feature represents $F_t$ and $F_x$. For patch-mixed image, the multi-level patch-mixed strategy are designed to obtain $F_{\hat{x_j}}$. We calculate the text-to-image similarity and select image samples with high similarity and high PCS scores. In the ASF module, we design the rule (a) and (b) to calculate the mIoU between synthetic annotations and the response of CLIP. The mIoU scores are sorted and we select the Top $n$ annotation samples. (3) The selected dataset remaining 26k images-annotation pairs are used to train a segmenter.}
\label{method}
\end{figure*}

\section{Related Work}


\subsection{Fully Supervised Semantic Segmentation}

Semantic segmentation is a pixel-level image analysis technique aimed at identifying and differentiating objects of all classes in an image, precisely locating them within the image. Currently, mainstream semantic segmentation methods primarily fall into two categories: fully convolutional neural networks (FCN)~\cite{long2015fully} and Transformer-based approaches. Among them, FCN includes models such as U-Net~\cite{ronneberger2015u}, SegNet~\cite{badrinarayanan2017segnet} and the DeepLab series~\cite{liang2015semantic, chen2017deeplab, chen2017rethinking, chen2018encoder}, and Mask2Former~\cite{cheng2022masked} for Transformer-based methods. All these methods are trained with images collected from real-world with manual pixel-level annotations, while in this work, we focus on synthetic image-annotation pairs.


\subsection{Synthetic Data for Semantic Segmentation}

Semantic segmentation of synthetic data is predominantly executed through image-to-mask and mask-to-image methods. For image-to-mask, 
DiffuMask~\cite{wu2023diffumask} exploits cross-attention maps between text and images and trains an Affinity Net to generate pixel-level annotations. Dataset-Diffusion~\cite{nguyen2024dataset} removes the Affinity Net and rewrites prompts by LLM, which only utilizes the diffusion model to generate accurate segmentation masks for synthetic images. For mask-to-image, FreestyleNet~\cite{xue2023freestyle} introduces Rectified Cross-Attention (RCA), seamlessly integrating semantic masks into the image generation process. FreeMask~\cite{yang2024freemask} makes improvements based on FreestyleNet. It significantly improves the performance of semantic segmenter by filtering noise and prioritizing the sampling of hard-to-learn masks. 

\section{Method}
\subsection{Problem Setting}
We first utilize LLM to generate captions for real images. Then, images and their captions are input to the generative model~\cite{nguyen2024dataset} to synthetic the dataset $D=(X_m, Y_m)^M_{m=1}$, where $X_m$ is the image and $Y_m$ is the corresponding annotation. These images and annotations capture both the semantic and location information of the target classes $C=\{c_1, c_2, ..., c_L\}$, where $L$ represents the number of classes. Our objective is to select high-quality samples from $D$ to perform a reliable synthetic dataset $D_{f}=(X_m,Y_m)^{{M}_{f}}_{m=1}$, where ${M}_{f}$ represents the number of selected samples. Finally, we can train a segmenter with ${D}_{f}$.


\subsection{Overview}
The overall framework of our approach is shown in Fig.~\ref{method}, which can be divided into the following steps:


 
 

\begin{enumerate}[1)]
 \item We generate massive synthetic images with corresponding annotations following Dataset-Diffusion~\cite{nguyen2024dataset}. The original synthetic images are input to our PCS module to evaluate their quality by comparing the similarity and our designed PCS scores. We select high-fidelity images that exhibit both high similarity and PCS scores.

 \item Meanwhile, the reference annotation for each image is generated using softmax Class Activation Maps~\cite{lin2023clip}. Both the reference annotations and the synthetic annotations are input into the ASF module to assess their quality. We select reliable annotations based on our designed mIoU-based rule.
 \item We combine the selected images from the PCS module with annotations from the ASF module to create a reliable dataset for training a segmenter.

\end{enumerate}
 
\subsection{Synthetic Dataset Generation}


Recent studies indicate that high-quality and diverse results can be synthesized by training large-scale text-to-image diffusion models. 
Motivated by this, we adopt the most recent work Dataset-Diffusion~\cite{nguyen2024dataset}, to synthesize additional training image-annotation pairs based on the provided real datasets. 
Specifically, for each real image, we generate $K$ captions through the LLM, such as ChatGPT~\cite{openai2023gpt}, then captions are input to the diffusion model to obtain $K$ synthetic images. Meanwhile, the corresponding annotations are generated using the attention maps from the diffusion model. These synthetic samples (image-annotation pairs) build an initial dataset.



\subsection{Perturbation-based CLIP Similarity (PCS)}
A straightforward way to evaluate image quality is to use the CLIP model to calculate the cosine similarity between the text prompt features and the corresponding image features directly. Specifically, the text prompt $T$ and the image $X$ are input into the text encoder ${f}_{\theta}$ and image encoder ${g}_{\theta}$, respectively, obtaining two feature representations:
\begin{equation}
   F_t={f}_{\theta }(T),F_x={g}_{\theta }(X) , \label{eq:1}
\end{equation} 
where $F_t \in \mathbb{R}^{1 \times d}$ and $F_x \in \mathbb{R}^{1\times d}$ are the feature representations of the text prompt $T$ and the image $X$, respectively, where $d$ is the channel number. 

The common text-to-image similarity ${S}_{x} \in [0, 1]$ is calculated as:

\begin{equation}
     {S}_{x}=\frac{{F}_{t} {F}_{x}^{\top}}{\parallel {F}_{t} \parallel \cdot \parallel {F}_{x} \parallel }, \label{eq:2}
\end{equation}
where $\top$ represents the matrix transpose, and $\parallel \cdot \parallel$ represents the $l_{2}$ norm. Our experiments indicate that synthetic data usually have high text-to-image similarity, so only relying on Eq.~(\ref{eq:2}) cannot evaluate the quality of synthetic data.

To address this problem, we propose Perturbation-based CLIP Similarity (PCS). Perturbing in synthetic images can be achieved through image transformations such as pixel-mixed, patch-mixed, or random occlusion. Each strategy possesses distinct characteristics. With pixel-mixed, the mean color of the image is maintained, but it becomes difficult to discern the object feature. Patch-mixed disrupts the shape of the object in the image but preserves local features through the patches. Random occlusion allows for the preservation of partial information. But, when the object is not exceptionally large, it may fully occlude the object~\cite{lee2024entropy}. In our PCS module, we need both the object features in images before and after perturbation. Therefore, we apply the patch-mixed strategy in our method, which divides an image into patches with different scales and then mixes the order of patches as new images to keep semantic integrity.

\begin{figure}[!t]
\centerline{\includegraphics[width=1.0\columnwidth]{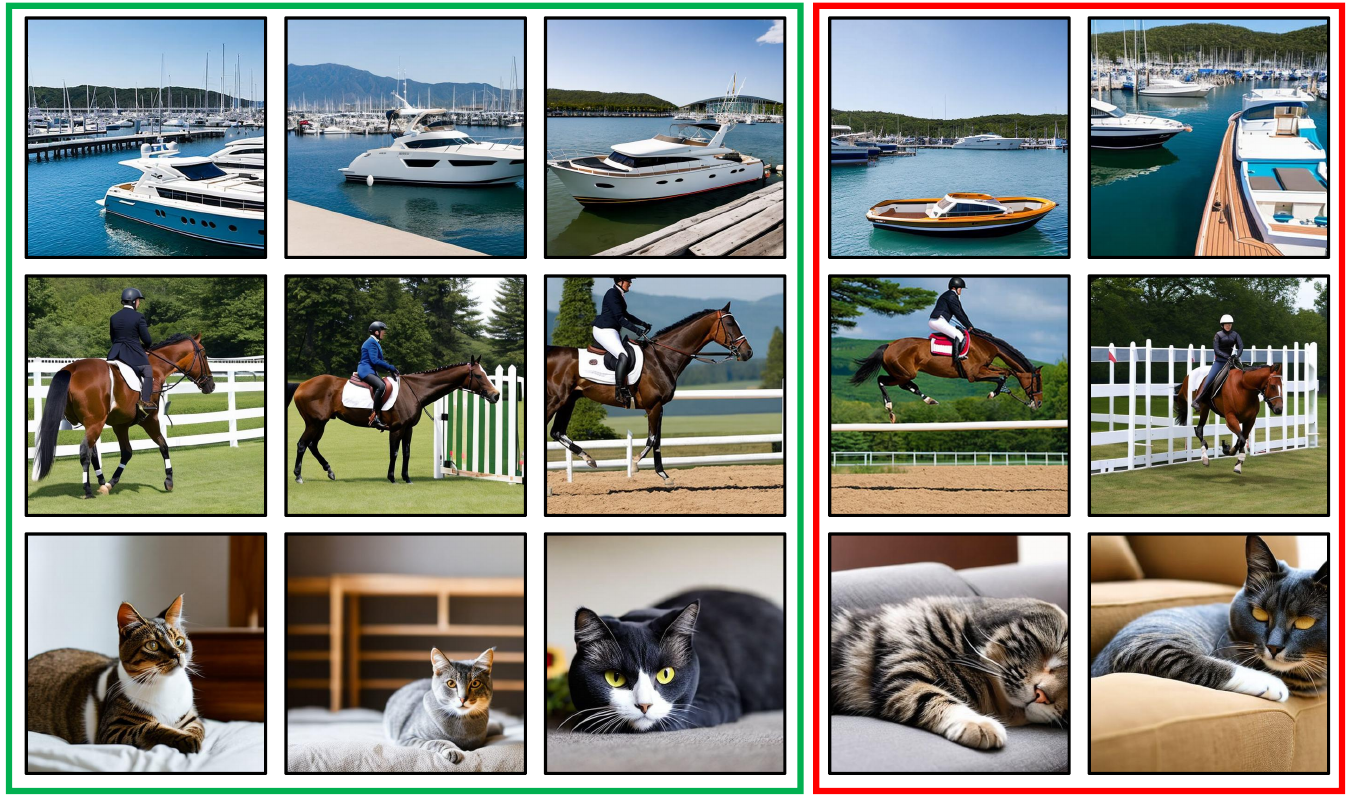}}
\caption{Visualization of selected images. High-quality images are with \textcolor{green}{green box} and low-quality with \textcolor{red}{red box}.}
\label{image}
\end{figure}



Specifically, 
given an original image $X\in R^{H\times W\times 3}$, we design a multi-level patch-mixed strategy, which sets $N_s$ patch scales and produces $N_o$ different patch orders for each scale to mitigate the influence of the random. If we only perform once patch-mixed on the image, different patch sizes and orders will lead to unstable similarity and affect the reproducibility of the experiment. Finally, we can generate $N_s\times N_o$ new mixed images. In detail, to generate the $j$-th mixed image, suppose the original image is divided into $n$ patches and the produced random patch order is $\left\{ j_{1}, j_{2}, ..., j_{n} \right\}$, for example, 4 patches with the order $\left\{ 2, 1, 4, 3 \right\}$, then the mixed image is defined as follows:
\begin{equation}
\begin{aligned}
X_j=\left\{patch_{j_{1}},patch_{j_{2}},\dots,patch_{j_{n}} \right\}
\end{aligned},
\end{equation}
where $patch_{j_{*}}$ means the $j_{*}$ patch in the original image. $X_j$ is the $j$-th mixed image, and $j \in \left\{1,2,\dots, N_s \times N_o\right\}$. For each $X_j$, we obtain the visual feature $F_{\hat{x}_j}={g}_{\theta }(X_j)$ through Eq.~(\ref{eq:1}). Then, we compute the similarity between the mixed image and the text as follows:


\begin{equation}
     {S}_{\hat{x}_j}=\frac{{F}_{t} F_{\hat{x}_j}^{\top}}{\parallel {F}_{t} \parallel \cdot \parallel F_{\hat{x}_j} \parallel } .\label{eq:4}
\end{equation}


Finally, we average all similarities computed from  $N_s\times N_o$ patch-mixed images as follows:
\begin{equation}
{S}_{\hat{x}}=\frac{1}{N_s \times N_o}\displaystyle\sum_{j=1}^{N_s\times N_o}\displaystyle{S}_{\hat{x}_j} ,
    \label{eq:5}
\end{equation}
${S}_{\hat{x}}$ is the averaged similarity, it integrates representative information from multiple mixed images, making the similarity more reliable. With the averaged similarity ${S}_{\hat{x}}$ and text-to-image similarity $S_x$, we design Perturbation-based CLIP Similarity to select high-quality samples:

\begin{equation}
\begin{gathered}
G(x)=\left \{ x\mid S_x> {\tau }_{s},PCS(x,\hat{x})> {\tau }_{PCS}\right \}, \\
    \label{eq:6}
\end{gathered}
\end{equation}
where
\begin{equation}
PCS(x,\hat{x})=S_x-{S}_{\hat{x}},
\label{eq:pcs score}
\end{equation}
In Eq.~(\ref{eq:6}), $G(x)$ is the selected image set that includes images maintaining both high text-to-image similarity and high PCS score, ${\tau }_{s}$ and ${\tau }_{PCS}$ are thresholds. Eq.~(\ref{eq:pcs score}) computes the PCS score for the images. The selected images have two advantages: a) rich semantics of objects, the condition $S_x> {\tau }_{s}$ ensures a strong correlation between the semantics of the object and text prompt; b) Fit the real distribution, $PCS(x,\hat{x})> {\tau }_{PCS}$ ensures synthetic images fit in a similar distribution with real images and contain representative information. As shown in Fig.~\ref{image}, the selected images are highlighted with the green box. We observe that the images in the green box are visually pleasant, objects in such images contain relatively rich semantics. Yet, the images in the red box are less realistic, and our PCS can help discard these low-quality samples.



\begin{figure}[!t]
\centerline{\includegraphics[width=1.0\columnwidth]{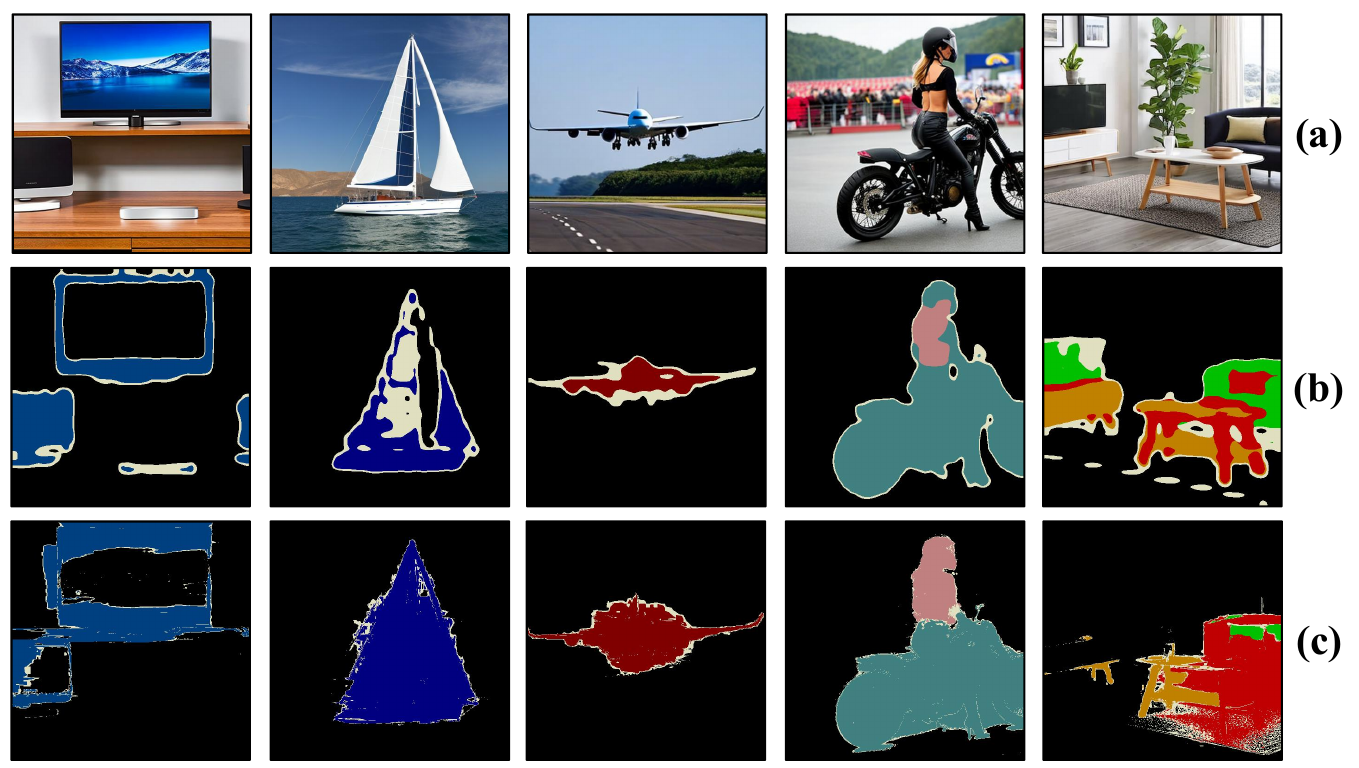}}
\caption{Visualization of low-quality annotation examples. (a) Synthetic Images. (b) Synthetic Annotations. (c) Reference Annotations from CLIP.}
\label{annotation}
\end{figure}

\subsection{Class-balance Annotation Similarity Filter (ASF)}
Due to the limitations of the generative models, the synthetic annotations unavoidably contain multiple objects that are relatively inaccurate~\cite{nguyen2024dataset}. As shown in Fig.~\ref{annotation}(b), there are some low-quality annotation examples. To remove these samples, we propose a class-balance Annotation Similarity Filter (ASF). The most important metric for evaluating annotations is mIoU which measures the degree of overlap between two segmentation regions, with higher values indicating better segmentation accuracy. Owing to the synthetic images not being manually annotated, we generate a set of reference annotations for the selected images in previous section from CILP, following~\cite{lin2023clip}.
 

Considering the impact of the class, our ASF module follows two rules: a) The quality of annotations containing fewer classes is generally higher. To balance the number of classes in an annotation, we should group the dataset according to the class numbers in an annotation; b) The quality of annotations among the classes is unbalanced, \emph{e.g.}, images of the monitor have coarse annotations while images of sheep all have accurate annotations. Thus, we should group the dataset according to the object class in an annotation to prevent the classes with all low-quality annotations from being completely removed in the selection process.
We define ${Y}^{num}$ and  ${Y}^{cls}$ to represent two grouped annotations based on the above two rules, respectively:

\begin{equation}
{Y}^{num}=\left \{ Y^{num}_{1},Y^{num}_{2},\dots ,Y^{num}_{q},\dots,Y^{num}_Q \right \},\\
\label{eq:8}
\end{equation}

\begin{equation}
{Y}^{cls}=\left \{ Y^{cls}_{1},Y^{cls}_{2},\dots ,Y^{cls}_{r},\dots, Y^{cls}_R \right \}.
\label{eq:9}
\end{equation}

In Eq.~(\ref{eq:8}), $Y^{num}_{q}$ is the $q$-th grouped annotation subset in $Y^{num}$, where $q \in \left\{ 1, 2, .., Q \right\}$. $Q$ represents the maximum number of simultaneous classes among all the images. For example, $Q=6$ means up to six classes appear among all images. $Y^{cls}_{r}$ is the $r$-th subset in $Y^{cls}$ and $r \in \left\{ 1, 2, .., R \right\}$.  $R$ represents the maximum index of classes in the dataset, \eg, $R=20$ means the 20th class in the dataset. Then for each subset, 
we calculate the mIoU score as follows:
\begin{equation}
{U}^{num}_q=\left \{ U^{num}_{q_1},U^{num}_{q_2},\dots ,U^{num}_{q_\text{max}} \right \},
\label{eq:10}
\end{equation}

\begin{equation}
{U}^{cls}_r=\left \{ U^{cls}_{r_1},U^{cls}_{r_2},\dots ,U^{cls}_{r_\text{max}} \right \},
\label{eq:11}
\end{equation}
where $U^{num}_{q_i}$ is the mIOU score for the $i$-th image in the $Y_q^{num}$. Similarly, $U^{cls}_{r_j}$ is the mIoU for the $j$-th image in the $Y_r^{cls}$. $q_\text{max}$ and $r_\text{max}$ are the maximum numbers of the corresponding grouped annotations. The ${U}^{num}_{q_i}$ and ${U}^{cls}_{r_j}$ are computed as follows:
\begin{equation}
\resizebox{0.88\columnwidth}{!}{$
{U}^{num}_{q_i}=\text{mIoU}(Y_{q_i}^{num}, \hat{Y}_{q_i}^{num}), q_i \in \left\{ q_1, q_2, ..., q_\text{max} \right\},\\
$}\label{eq:12}
\end{equation}

\begin{equation}
\resizebox{0.88\columnwidth}{!}{$
{U}^{cls}_{r_j}=\text{mIoU}(Y_{r_j}^{cls}, \hat{Y}_{r_j}^{cls}), r_j \in \left\{ r_1, r_2, ..., r_\text{max} \right\},\\
\label{eq:13}
$}
\end{equation}
where $Y_{q_i}^{num}$ and $Y_{r_j}^{cls}$ are synthetic annotations, $\hat{Y}_{q_i}^{num}$ and $\hat{Y}_{r_j}^{cls}$ are reference annotations generated from CLIP. A higher mIoU score means the synthetic annotation is more reliable. Note that each element index in $Y^{num}_q$ and $U^{num}_q$ is a one-to-one correspondence. Hence, we can use the mIoU set, $U^{num}_q$, to select the annotations from $Y^{num}_q$, \emph{i.e.}, annotations with Top $n$ mIoU scores in each grouped annotation remain to build a new subset, and all other annotations are removed, as follows:
\begin{equation}
\begin{gathered}
\text{Top}_n(Y^{num}_q) = \left\{ Y^{num}_{q_{k_1}}, Y^{num}_{q_{k2}}, \ldots, Y^{num}_{q_{k_n}} \right\}, \\
\text{where} \quad U^{num}_{q_{k_1}} \geq U^{num}_{q_{k_2}} \geq \cdots \geq U^{num}_{q_{k_n}}.
\label{eq:14}
\end{gathered}
\end{equation}

For annotation set $Y^{cls}_r$, we obtain $\text{Top}_n(Y^{cls}_r)$ following the above process. The reliable annotation set is selected as:
\begin{equation}
\begin{gathered}
H(Y)=\left \{ Y \mid Y \in \text{Top}_n(Y^{num}_q) \cup \text{Top}_n(Y^{cls}_r) \right\}
\label{eq:15}
\end{gathered}
\end{equation}
where $H(Y)$ represents the selected annotation set which is the union of $\text{Top}_n(Y^{num}_q)$ and $\text{Top}_n(Y^{cls}_r)$. 

Finally, we combine the two modules to build a reliable synthetic training dataset consisting of high-fidelity images with their corresponding high-quality pixel-level semantic annotations. 




\section{Experiments}
\begin{table*}[h] \footnotesize
  \centering  
  \begin{threeparttable}
    \begin{tabular}{>{\centering\arraybackslash}p{6.0cm}|
                    >{\centering\arraybackslash}p{4.5cm}|
                    >{\centering\arraybackslash}p{1.6cm}
                    >{\centering\arraybackslash}p{1.5cm}
                    >{\arraybackslash}p{1.5cm}} 
    \toprule  
    Method&Segmenter&Backbone&Data Size&mIoU (\%)\cr
    \midrule
    VOC’s training~\cite{everingham2015pascal}&\multirow{6}{*}{DeepLabV3~\cite{chen2017rethinking}}&\multirow{3}{*}{ResNet50}&11.5k&77.4\cr
    Dataset Diffusion~\cite{nguyen2024dataset}&&&40k&58.1*\cr
    SDS-VOC (Ours)&&&26k&\textbf{60.4}\cr
    \cline{1-1}\cline{3-5}
    VOC’s training~\cite{everingham2015pascal}&&\multirow{3}{*}{Resnet101}&11.5k&79.9\cr
    Dataset Diffusion~\cite{nguyen2024dataset}&&&40k&56.9*\cr
    SDS-VOC (Ours)&&&26k&\textbf{59.1}\cr
    \midrule
    VOC’s training~\cite{everingham2015pascal}&\multirow{4}{*}{Mask2Former~\cite{cheng2022masked}}&\multirow{4}{*}{Resnet50}&11.5k&77.3\cr
    DiffuMask~\cite{wu2023diffumask}&&&60k&57.4\cr
    Dataset Diffusion~\cite{nguyen2024dataset}&&&40k&57.8*\cr
    SDS-VOC (Ours)&&&26k&\textbf{59.8}\cr
    \midrule
    Dataset Diffusion~\cite{nguyen2024dataset}&\multirow{2}{*}{CDL~\cite{zhang2023credible}}&\multirow{2}{*}{Resnet101}&40k&59.6*\cr
    SDS-VOC (Ours)&&&26k&\textbf{62.5}\cr
    \bottomrule  
    \end{tabular}  
    \caption{Comparisons with other methods on the PASCAL VOC 2012 dataset. * means our reproduced results.}
    \label{tab:voc}
    \end{threeparttable}  
\end{table*}

\begin{table*}[h] \footnotesize
  \centering  
  \begin{threeparttable}
    \begin{tabular}{>{\centering\arraybackslash}p{6.0cm}|
                    >{\centering\arraybackslash}p{4.5cm}|
                    >{\centering\arraybackslash}p{1.6cm}
                    >{\centering\arraybackslash}p{1.5cm}
                    >{\arraybackslash}p{1.5cm}} 
    \toprule  
    Method&Segmenter&Backbone&Data Size&mIoU (\%)\cr
    \midrule
    COCO’s training~\cite{lin2014microsoft}&\multirow{6}{*}{DeepLabV3~\cite{chen2017rethinking}}&\multirow{3}{*}{ResNet50}&118k&48.9\cr
    Dataset Diffusion~\cite{nguyen2024dataset}&&&80k&28.7*\cr
    SDS-COCO (Ours)&&&50k&\textbf{31.0}\cr
    \cline{1-1}\cline{3-5}
    COCO’s training~\cite{lin2014microsoft}&&\multirow{3}{*}{Resnet101}&118k&54.9\cr
    Dataset Diffusion~\cite{nguyen2024dataset}&&&80k&29.2*\cr
    SDS-COCO (Ours)&&&50k&\textbf{31.8}\cr
    \midrule
    Dataset Diffusion~\cite{nguyen2024dataset}&\multirow{2}{*}{CDL~\cite{zhang2023credible}}&\multirow{2}{*}{Resnet101}&80k&30.3*\cr
    SDS-COCO (Ours)&&&50k&\textbf{33.4}\cr
    \bottomrule  
    \end{tabular}  
    \caption{Comparisons with other methods on the MS COCO 2017 dataset. * means our reproduced results.} 
    \label{tab:coco}
    \end{threeparttable}  
\end{table*}

\subsection{Dataset and Evaluation Metrics}

We evaluate our method on PASCAL VOC 2012~\cite{everingham2015pascal} and MS COCO 2017~\cite{lin2014microsoft}.

\textbf{PASCAL VOC 2012} has 20 object classes and 1 background class, which is augmented by SBD~\cite{hariharan2011semantic} to obtain 10,584 training, 1,449 validation, and 1,456 test images. For the synthetic dataset, we follow Dataset Diffusion~\cite{nguyen2024dataset} to produce 40k image-annotation pairs as the initial dataset. After applying SDS on the initial dataset, 26k high-quality samples are selected to form the SDS-VOC dataset.


\textbf{MS COCO 2017} contains 80 object classes and 1 background class with 118k training and 5k validation images. Similarly, we follow Dataset Diffusion to obtain 80k synthetic image-annotation pairs and then use our method to get 50k high-quality samples to form the SDS-COCO dataset.

The mean Intersection over Union (mIoU) is used as the evaluation metric.

\subsection{Implementation Details}


\textbf{Model settings:} We employ the CLIP pre-trained model ViT-B-16 with our method to select high-quality samples. We involve three segmenters for evaluation: DeepLabv3~\cite{chen2017deeplab}, Mask2Former~\cite{cheng2022masked}, and CDL~\cite{zhang2023credible}. All segmenters follow the default settings in the original paper. 

\textbf{Hyperparatemers:} In Perturbation-based CLIP Similarity (PCS), $N_s\in \left \{ 8,16,32 \right \} $ represents the patch scale. $N_o$ is the number of patch orders, we set $N_o=3$. In Eq.~\ref{eq:6}, thresholds ${\tau }_{s}$ and ${\tau }_{PCS}$ are set to 0.8 and 0.1, respectively. In class-balance Annotation Similarity Filter (ASF),
we set $n$ to 0.6 $\times$ the number of samples within groups in Eq.~(\ref{eq:15}), \emph{i.e.}, top 60\% samples are selected. All experiments are running on NVIDIA RTX 3090 GPUs.


\subsection{Performance Comparison}
Table~\ref{tab:voc} presents the evaluation results of three segmenters. The datasets used in Table~\ref{tab:voc} are divided into three types, (1) real dataset; (2) initial datasets where samples are generated by diffusion models (Dataset Diffusion); (3) SDS dataset, comprising only high-quality samples selected from the initial dataset. From comparison, our method achieves 62.5\% mIoU when compared to the Dataset diffusion of 59.6\% mIoU. Further, ours outperforms DiffuMask by 2.4\% mIoU using the same ResNet50 backbone. The results reveal the diffusion model produces many noise samples which hinder the semantic segmentation, with our method, only high-quality samples remain, and achieve better performance.

\begin{figure}[!t]
\centerline{\includegraphics[width=1.0\columnwidth]{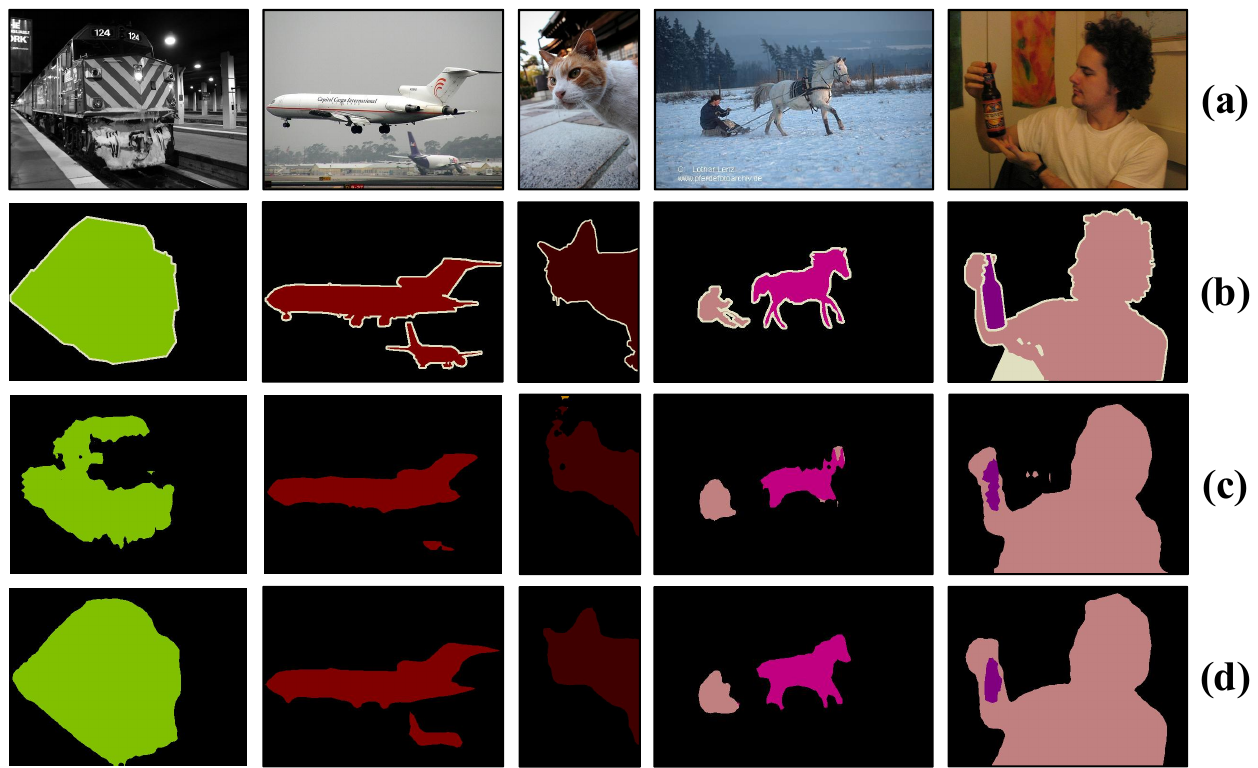}}
\caption{Segmentation results on the PASCAL VOC 2012 validation set. (a) Images. (b) Ground-truth annotation. (c) Predicted annotation from Dataset Diffusion. (d) Our predicted annotation.} 
\label{pre}
\end{figure}

Fig.~\ref{pre} shows our predicted annotation results on the validation set of VOC, which overall align with the ground truth. 

\textbf{MS COCO 2017:} Table~\ref{tab:coco} shows the performance on two segmenter. The dataset settings are the same as above. Our method achieves a promising result of 33.4\% mIoU compared to 30.3\% mIoU of the Dataset Diffusion. 


\subsection{Ablation Studies}
We select DeepLabV3~\cite{chen2017rethinking} with ResNet50 as the segmenter. All experiments are conducted on the PASCAL VOC 2012 dataset unless otherwise stated.


\textbf{Effectiveness of Each Module in SDS:}
Table~\ref{tab:module} demonstrates the effectiveness of each module in SDS by progressively adding PCS and ASF. It can be seen both PCS and ASF bring increased performance. By combining PCS and ASF to select a dataset for training segmenter, the segmenter has the best result, reaching 60.4\% mIoU. PCS module selects high-quality images and the ASF module selects high-quality annotations, which are complementary and consistent with our design targets. 

\begin{figure}[!t]
\centerline{\includegraphics[width=1.0\columnwidth]{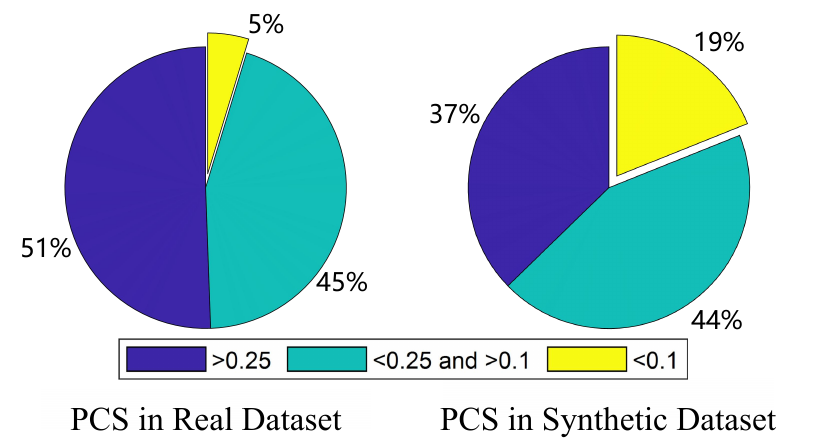}}
\caption{A significant difference in PCS scores between the real dataset and the synthetic dataset.}
\label{pie}
\end{figure}

Moreover, we make a comparison of our proposed PCS score between the real dataset and the initial synthetic dataset. As shown in Fig.~\ref{pie}, the proportion of low PCS scores in the synthetic dataset is much higher than that in the real dataset, which indicates that using PCS is an effective method to evaluate the image quality.

\begin{table}[!t] \footnotesize
  \centering  
  \begin{threeparttable}
    \begin{tabular}{>{\centering\arraybackslash}p{1.2cm}
                    >{\centering\arraybackslash}p{1.2cm}
                    >{\centering\arraybackslash}p{1.2cm}|
                    >{\centering\arraybackslash}p{1.5cm}}
    \toprule  
    base&PCS&ASF&mIoU (\%)\cr
    \midrule
    \checkmark     &     &     &  58.1\cr
    \checkmark     &\checkmark     & &   59.5\cr
    \checkmark     &\checkmark     &\checkmark     &  \textbf{60.4}\cr
    \bottomrule  
    \end{tabular}  
    \caption{Effectiveness of PCS and ASF modules. ``base" means the synthetic dataset of Dataset Diffusion.} 
    \label{tab:module}
    \end{threeparttable}  
\end{table}


\textbf{Multi-level Patch-mixed Strategy:}
Table~\ref{tab:patch} illustrates the influence of $N_s$ and $N_o$ in Multi-level Patch-mixed Strategy. $N_s$ represents the scales of patches in the image and $N_o$ represents the number of orders for each scale. It can be observed that the optimal result is achieved when we take average of $N_s$ patch scales and set the $N_o=3$. 
\begin{table}[!t] \footnotesize
  \centering  
  \begin{threeparttable}
    \begin{tabular}{>{\centering\arraybackslash}p{1.6cm}
                    >{\centering\arraybackslash}p{1.0cm}|
                    >{\centering\arraybackslash}p{1.5cm}}
    \toprule  
    $N_s$ & $N_o$ & mIoU (\%)\cr
    \midrule
       \{8\} &   3  &       59.9\cr
       \{16\} &   3  &       60.2\cr
       \{32\} &    3 &       59.5\cr
       \{8, 16, 32\} &  3   &       \textbf{60.4}\cr
    \bottomrule  
    \end{tabular}  
    \caption{Ablation studies of multi-level patch-mixed strategy.} 
    \label{tab:patch}
    \end{threeparttable}  
\end{table}


\textbf{Effectiveness of Rules in ASF Module:} Table~\ref{tab:ASF} demonstrates the effectiveness of each rule in the ASF module. We observe that both Rule (a) and Rule (b) bring increased performance, which indicates the necessity of our class-balance strategy.

\begin{table}[!t] \footnotesize
  \centering  
  \begin{threeparttable}
    \begin{tabular}{>{\centering\arraybackslash}p{1.2cm}
                    >{\centering\arraybackslash}p{1.2cm}
                    >{\centering\arraybackslash}p{1.2cm}|
                    >{\centering\arraybackslash}p{1.3cm}}
    \toprule  
    Direct&Rule (a)&Rule (b)&mIoU (\%)\cr
    \midrule
    \checkmark     &     &     &  57.5\cr
         &\checkmark     &     &  59.2\cr
        &    &\checkmark     &  59.6\cr
        & \checkmark   &\checkmark     &  \textbf{60.4}\cr
    \bottomrule  
    \end{tabular}  
    \caption{Effectiveness of rules in ASF module. ``Direct" means calculating the mIoU in all synthetic annotations and taking annotations with Top $n$ mIoU scores.} 
    \label{tab:ASF}
    \end{threeparttable}  
\end{table}

\textbf{Data Size and the Performance of Segmenter:} Table~\ref{tab:k} illustrates the relationship between data size and the segmenter performance. Our method can reduce the synthetic training dataset by half but generate a 2.3\% mIoU increase.
\begin{table}[!t] \footnotesize
  \centering  
  \begin{threeparttable}
    \begin{tabular}{>{\centering\arraybackslash}p{1.8cm}|
                    >{\centering\arraybackslash}p{1.0cm}
                    >{\centering\arraybackslash}p{1.0cm}
                    >{\centering\arraybackslash}p{1.0cm}
                    >{\centering\arraybackslash}p{1.0cm}}
    \toprule  
    Data Size& 40k& 30k & 26k& 15k\cr
    \midrule
    mIoU (\%)& 58.1& 59.7 & \textbf{60.4}& 58.0\cr
    \bottomrule  
    \end{tabular}  
    \caption{The relationship between data size and the performance of segmenter.} 
    \label{tab:k}
    \end{threeparttable}  
\end{table}

\section{Conclusion}
In this work, we propose a training-free Synthetic Data Selection (SDS) method with CLIP to select high-quality samples from synthetic dataset. To achieve this, we design two novel modules: the PCS module, which introduces Perturbation in images and selects high-quality images without incorrect object relationships, and the ASF module, which applies a class-balance strategy and selects high-quality annotations based on mIoU scores.
With our two processing strategies, the segmenter trained on the selected dataset achieves a better performance. 

\section{Acknowledgments}

This work was supported by the National Natural Science Foundation of China (Grant No.62301613, No.62301451), the Taishan Scholar Program of Shandong (No. tsqn202306130), the Shandong Natural Science Foundation (Grant No. ZR2023QF046), Independent Innovation Research Project of China University of Petroleum (East China) (No.22CX06060A).

\bibliography{aaai25}

\end{document}